\documentclass[10pt,twocolumn,letterpaper]{article}
\usepackage[dvipsnames]{xcolor}
\usepackage{multirow}
\usepackage{algorithm}
\usepackage{algpseudocode}
\usepackage{amsmath}
\usepackage[pagenumbers]{cvpr} 
\usepackage{makecell}
\definecolor{cvprblue}{rgb}{0.21,0.49,0.74}
\usepackage[pagebackref,breaklinks,colorlinks,citecolor=cvprblue]{hyperref}
\title{Generalized Category Discovery in Semantic Segmentation}
\author{%
  Zhengyuan Peng$^{1,*}$ \quad Qijian Tian$^{1,*}$ \quad Jianqing Xu$^{3}$ \quad  Yizhang Jin$^{1}$ \\
  Xuequan Lu$^{4}$  \quad Xin Tan$^{2}$ \quad Yuan Xie$^{2}$ \quad Lizhuang Ma$^{1,2}$\\
  $^1$ Shanghai Jiao Tong University 
  $^2$ East China Normal University \\
  $^3$ Youtu Lab, Tencent 
  $^4$ La Trobe University \\
  pengzhengyuan@sjtu.edu.cn \\
}

\begin{document}
\maketitle

\newcommand{\jianqing}[1]{\textcolor{red}{[#1]}}

\begin{abstract}
This paper explores a novel setting called Generalized Category Discovery in Semantic Segmentation (GCDSS), aiming to segment unlabeled images given prior knowledge from a labeled set of base classes. The unlabeled images contain pixels of the base class or novel class. In contrast to Novel Category Discovery in Semantic Segmentation (NCDSS), there is no prerequisite for prior knowledge mandating the existence of at least one novel class in each unlabeled image. Besides, we broaden the segmentation scope beyond foreground objects to include the entire image. Existing NCDSS methods rely on the aforementioned priors, making them challenging to truly apply in real-world situations. We propose a straightforward yet effective framework that reinterprets the GCDSS challenge as a task of mask classification. Additionally, we construct a baseline method and introduce the Neighborhood Relations-Guided Mask Clustering Algorithm (NeRG-MaskCA) for mask categorization to address the fragmentation in semantic representation. A benchmark dataset, Cityscapes-GCD, derived from the Cityscapes dataset, is established to evaluate the GCDSS framework. Our method demonstrates the feasibility of the GCDSS problem and the potential for discovering and segmenting novel object classes in unlabeled images. We employ the generated pseudo-labels from our approach as ground truth to supervise the training of other models, thereby enabling them with the ability to segment novel classes. It paves the way for further research in generalized category discovery, broadening the horizons of semantic segmentation and its applications. For details, please visit https://github.com/JethroPeng/GCDSS
\end{abstract}

\section{Introduction}
\label{sec:intro}

Semantic segmentation is a fundamental task in computer vision that aims to partition an image into semantically meaningful regions, assigning each pixel to a specific class. It has been extensively studied, with many methods for various aspects of the problem. Most methods~\cite{FCN, DeepLab-v2, PSPNet, SETR, DeepLabv3+} predefine a fixed set of object classes, requiring corresponding labeled data for model training. However, real-world images may contain objects from novel classes, posing a challenge for traditional segmentation methods.\par

\begin{figure}[t]
\centering
\includegraphics[scale=0.43]{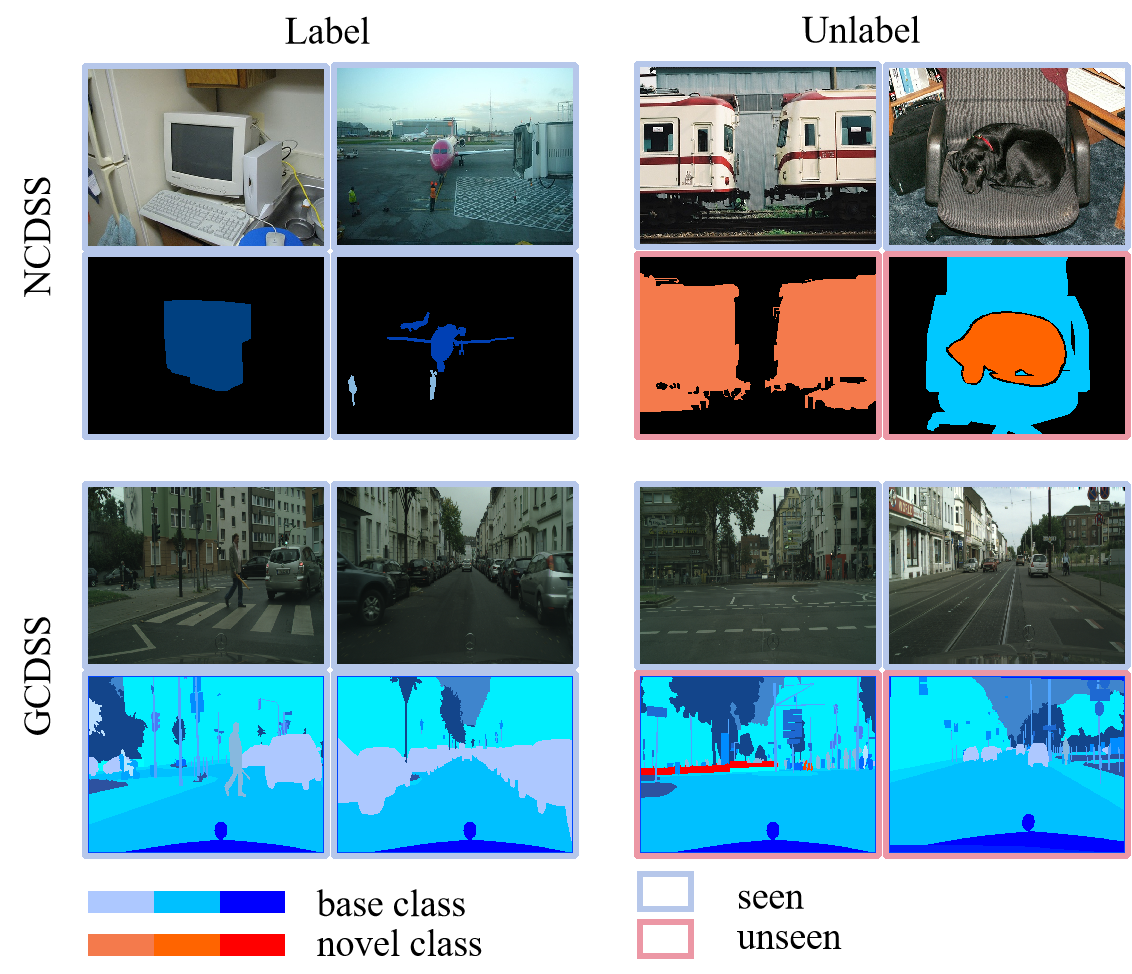}
\caption{\textbf{Illustration of Generalized Category Discovery in Semantic Segmentation (GCDSS).}  In contrast to Novel Class Discovery in Semantic Segmentation (NCDSS), GCDSS eliminates the prior knowledge for unlabeled images to contain pixels from novel classes (row 4, column 4). GCDSS broadens the segmentation scope beyond foreground objects. The proportion of the pixel area occupied by novel classes in GCDSS is typically low.}
\label{main}
\vspace{-0.1in}
\end{figure} 

\begin{figure}[t]
\centering
\includegraphics[scale=0.4]{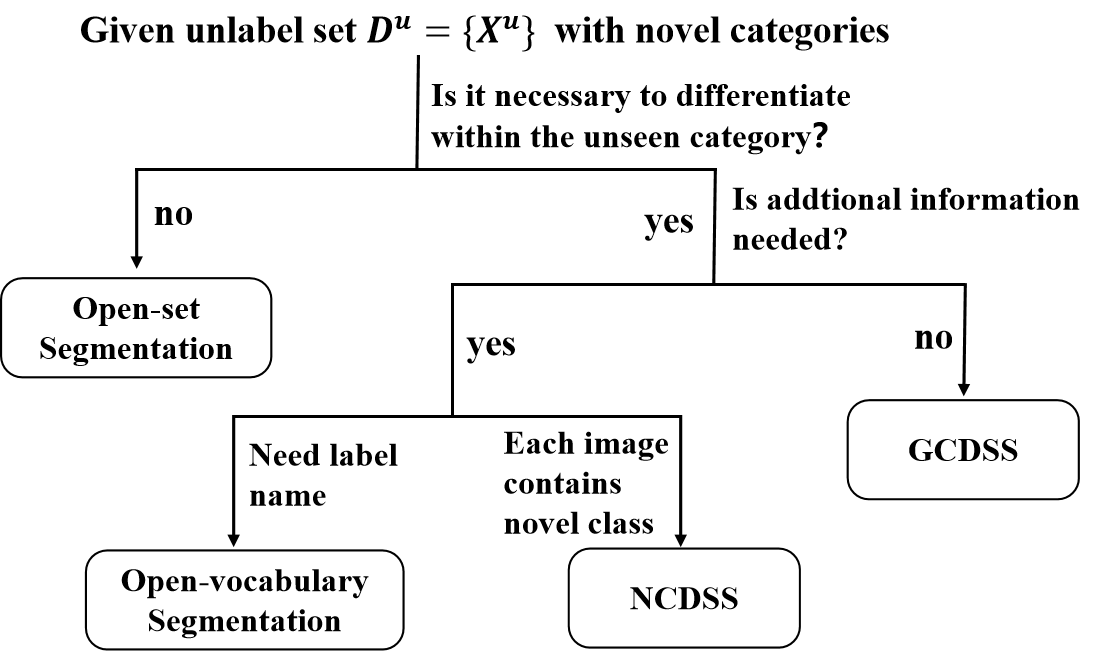}
\caption{\textbf{Comparison of different novel class segment settings.}}
\label{comp}
\vspace{-0.1in}
\end{figure}

Various settings address the challenge of novel classes in an unlabeled set. Open-set segmentation~\cite{xia2020synthesize,cen2021deep,hendrycks2016baseline,dehghani2023scaling} setting acknowledges the presence of novel classes but does not require them to be distinguished. Open-vocabulary segmentation~\cite{xu2022simple,ghiasi2022scaling,luo2023segclip,rao2022denseclip} requires names of novel classes. Novel Category Discovery in Semantic Segmentation (NCDSS)~\cite{NCDSS} hypothesizes that each image in the unlabeled set contains at least one object from novel classes (See Fig \ref{main}). Fig \ref{comp} compares these settings. Due to the requirements of prior information, they have limitations in practical scenarios.\par

This paper presents Generalized Category Discovery in Semantic Segmentation (GCDSS), inspired by Generalized Category Discovery (GCD) principles. GCD, introduced in \cite{GCD}, classifies an unlabeled set containing base and novel classes using only information from a labeled set of base classes. 
GCD is inspired by the way infants recognize the world. Using prior knowledge of familiar objects like chairs, infants cluster novel instances such as sofas into novel classes within their visual recognition system. GCD aspires for models to attain similar capabilities.\par

In the GCDSS setting, each labeled image only contains pixels from base classes, while the pixels of unlabeled images may belong to either base or novel classes. The objective is to segment the unlabeled set. GCDSS presents unique challenges compared to GCD, such as the finer granularity of the task, which increases the complexity and demands of the analysis. Furthermore, for the datasets under the setting of GCDSS, it is highly unlikely to find images containing only novel classes, as most images consist of a mix of base and novel classes. We provide a detailed discussion of these challenges and compare GCDSS with alternative settings in the following Preliminary Section (\cref{Preliminary}).\par

To address the GCDSS challenge, we propose a simple framework that transforms the GCDSS problem into a mask classification task. Our framework consists of three stages: mask generation, feature extraction, and clustering. In the mask generation stage, we create disjoint masks for each image, which serve as the basis for classification. During the feature extraction stage, we extract features from the generated masks. Finally, in the clustering stage, we cluster similar masks based on their features, aiming to discover and segment novel object classes. Furthermore, in accordance with practices, we provide a baseline method.\par

However, the introduction of mask generation inevitably leads to the challenge of discrete semantics, where a complete concept is divided into several masks with lower-level semantics. For instance, a person may be separated into distinct regions such as the head, torso, and legs. To address this challenge, we shift away from directly grouping these dispersed features and introduce the Neighborhood Relations-Guided Mask Clustering Algorithm (NeRG-MaskCA). NeRG-MaskCA is a novel approach comprising three steps: label propagation, structural completion, and clustering division. First, it assigns pseudo-labels to unlabeled masks by checking the labels of neighborhood masks within the feature space of labeled masks. The remaining masks not annotated are considered novel class masks with high confidence. Second, it eliminates the labels of masks in the neighborhood of novel class masks, ensuring that novel class masks retain a clustered structure in the feature space. The final step involves the clustering of these novel class masks. We leverage the novel class pseudo-labels generated through our approach as the designated ground truth to supervise other models. It enables conventional models to segment novel classees.\par

We present Cityscapes-GCD, a benchmark dataset designed for the GCDSS challenge. This dataset integrates a diverse mix of novel and base classes to form a comprehensive benchmark. Cityscapes-GCD is engineered to minimize the domain gap between the labeled and unlabeled sets, thereby sharpening the focus on the generalized category discovery. Additionally, there is an imbalance in pixel area between novel classes and base classes, simulating real-world scenarios.\par

In our evaluation metric, we introduce a rational approach to evaluating the performance of the GCDSS setting. Traditional GCD methods often use Hungarian matching for both base and novel classes. It may lead to unreasonable situations where the base class is not discovered. Our metric utilizes precise matching for base classes and incorporates a greedy matching technique for novel classes, ensuring a stringent and accurate performance assessment.\par

Our contributions can be summarized as follows:\par
$\bullet$ 
We build the GCDSS setting and benchmark, which extends traditional semantic segmentation to discover and segment objects from both base and novel classes, providing a more realistic setting for real-world applications.\par
$\bullet$ 
We present a straightforward yet efficient framework to transform the GCDSS problem into a mask classification task. We also introduce the Neighborhood Relations-Guided Mask Clustering Algorithm (NeRG-MaskCA), which facilitates the discovery of novel classes.\par
$\bullet$ 
Through extensive experiments, we prove the feasibility of addressing the GCDSS problem. Using our approach's pseudo-labels as ground truth, we enable other models to segment novel classes.it highlights our method's potential for discovering and segmenting novel classes.\par

\section{Related work}

\subsection{Semantic Segmentation}

Semantic segmentation achieves pixel-wise prediction through pixel-level supervised learning~\cite{FCN, PSPNet, DeepLab-v2, SETR, DeepLabv3+}. Besides per-pixel classification, mask classification is also commonly used for semantic segmentation. Mask R-CNN~\cite{MaskRCNN} uses a global classifier to classify mask proposals. DETR~\cite{DETR} proposes a Transformer~\cite{Transformer} design to handle thing-class segmentation. MaskFormer~\cite{MaskFormer} predicts a set of binary masks, and each of them is associated with a single class label. Recently, the large-scale segmentation model SAM~\cite{SAM} has demonstrated powerful segmentation capability. However, it tends to prioritize structural over semantic information. This limitation makes it less suitable for direct application in the GCDSS setting.\par

\subsection{Novel Class Discovery}

Novel Class Discovery aims to discover novel classes based on prior knowledge from base classes. The setting is formalized and solved by the two-stage method DTC in~\cite{DTC}. This method first extracts semantic representations with labeled images and fine-tunes the model with unlabeled image clustering. Following this work, some methods~\cite{UNO, NCL, OpenMix, Meta_Discovery, NCDwF, InterClassNCD, Bootstrap_Your_Own_Prior} utilize labeled images to discover novel classes in the NCD setting. In addition, NCDSS~\cite{NCDSS} extends the NCD to semantic segmentation. The following work~\cite{NCD3DPoint} addresses the NCD for 3D point cloud semantic segmentation. However, the NCD problem assumes that each unlabeled image in the unlabeled set must contain at least one novel class. It is a strong prior knowledge and is unrealistic since we often do not know whether novel classes exist in the unlabeled image. The pre-existing NCDSS model~\cite{NCDSS} relies on this prior design. Consequently, it leads to the non-transferability of the NCDSS model to our setting.

\subsection{Generalized Category Discovery}
Generalized Category Discovery~\cite{GCD} is also a setting that discovers novel classes by leveraging labeled data of base classes and unlabeled data. Different from NCD, it does not assume that images in the unlabeled set must contain novel classes. A simple yet effective semi-supervised k-means method~\cite{GCD} is first proposed to solve this problem. DCCL~\cite{DCCL} effectively improves clustering accuracy by alternately estimating underlying visual conceptions and learning conceptional representation. IGCD~\cite{IGCD} explores a category incremental learning setting to correctly categorize images from previously base categories, while also discovering novel ones. CLIP-GCD~\cite{CLIP-GCD} utilizes vision-language representations to solve the GCD problem. Some other methods~\cite{Parametric_GCD, Learning_Semi-supervised_GCD, MetaGCD} are also proposed for the GCD setting. However, current methods mainly focus on image classification. In our paper, we extend the GCD setting to semantic segmentation.

\begin{figure*}[h]
\centering
\includegraphics[width=1\textwidth]{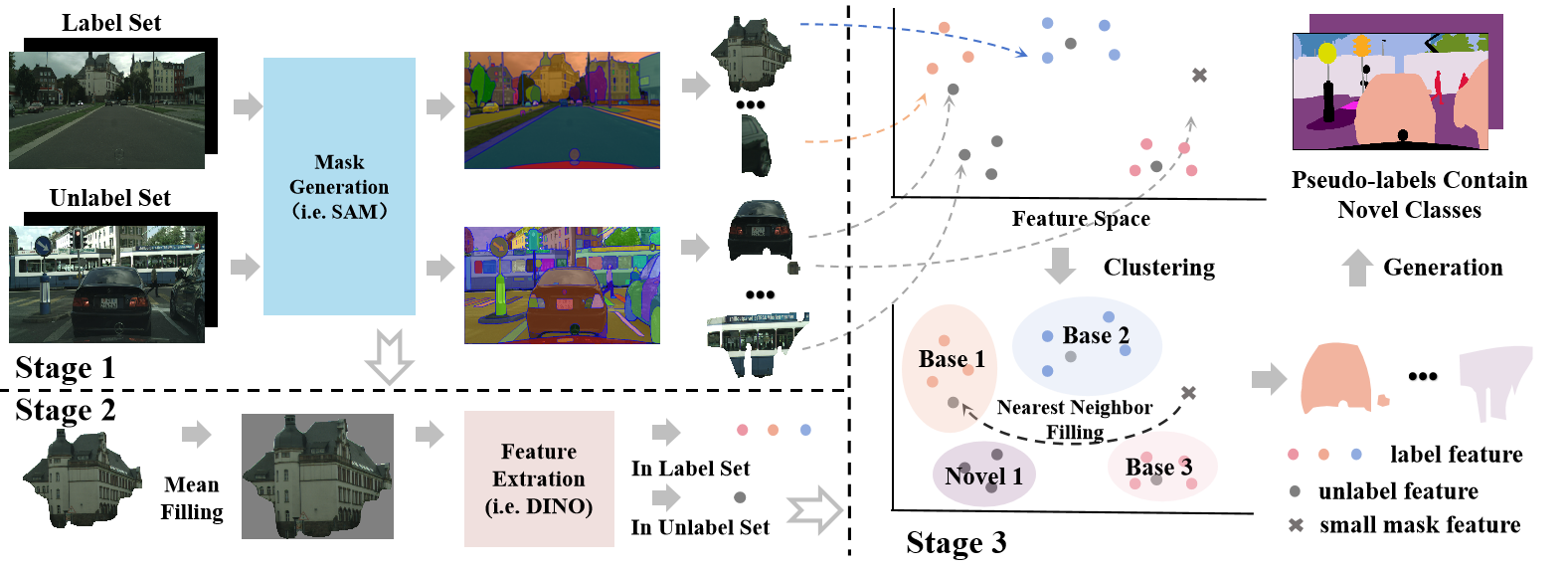}
\caption{\textbf{The baseline framework of GCDSS.} Our framework for GCDSS is divided into three key stages. 1. Mask Generation: the raw image serves as input to create distinct masks covering the entire image, transforming the semantic segmentation task into the mask classification task. 2. Feature Extraction: The masks are filled with the mean value to reduce the interference of background information. Features are extracted from the masks. 3. Clustering: Cluster labels are assigned to each mask based on their features. Small masks do not participate in clustering and maintain the same label as their nearest neighbor masks.}
\label{framework}
\end{figure*}

\section{Preliminary}\label{Preliminary}

\subsection{Problem Definition}

In the GCDSS setting, we involve two datasets: a labeled dataset and an unlabeled dataset. The labeled dataset, denoted as \( D_l = \{X_l, Y_l\} \), comprises images \( X_l \) and their corresponding labels \( Y_l \). This dataset includes a set of base classes, \( C_l \), containing \( N_l^c \) classes. Conversely, the unlabeled dataset, represented as \( D_u = \{X_u\} \), consists of images \( X_u \) that contain a set of classes, \( C_u \), encompassing \( N_u^c \) classes. The relationship between the class sets \( C_l \) and \( C_u \) is defined by \( C_l \subsetneqq C_u \). The objective of GCDSS is to segment pixels in the unlabeled images \( X_u \), which may include both base and novel classes, leveraging the knowledge from the labeled base classes in \( D_l \).

\subsection{GCD and GCDSS}

\textbf{Value.} GCDSS stands out from traditional GCD in several ways. Firstly, it not only identifies novel class objects in images but also accurately pinpoints their location and shape, providing more detailed and insightful information. GCDSS can segment multiple novel classes in the foreground or background. Besides, GCDSS benefits practice by reducing the expensive labeling costs associated with segmentation tasks, thus lightening the annotation burden.

\textbf{Challenge.} However, GCDSS also introduces certain challenges. One notable challenge is the finer granularity of the tasks it performs, which can make the analysis more complex and demanding. Moreover, for the datasets under the setting of GCDSS, it is almost impossible to find purely novel class images, as most images inevitably contain a mixture of base and novel classes.

\subsection{NCDSS and GCDSS}

\textbf{Value.} GCDSS introduces several differences over NCDSS. Firstly, it extends the segmentation scope from focusing on the foreground to encompassing the entire image. This broader scope allows for more comprehensive analysis by capturing both foreground and background elements. Additionally, GCDSS offers a higher degree of flexibility by not assuming the presence of novel classes in the unlabeled set as a prerequisite. These differences make GCDSS more suitable for real-world scenarios.

\textbf{Challenge.} GCDSS poses its unique challenges. NCDSS designs its models based on the prior assumption that each unlabeled image contains novel classes, while GCDSS has no such assumption. Additionally, during clustering in the unlabeled set, NCDSS has prior knowledge of the number of novel classes while GCDSS does not. The primary challenge in NCDSS is to achieve more accurate segmentation based on prior knowledge. In contrast, GCDSS may face challenges in distinguishing novel classes from base ones, and the difficulty lies in identifying all the novel classes comprehensively.

\section{Methods}

\subsection{Overview}

In this section, we propose a basic framework for addressing the challenging problem of Generalized Category Discovery. This framework is shown in Fig~\ref{framework}, which consists of three stages: mask generation (\cref{Mask Generation}), feature extraction (\cref{Feature Extraction}), and clustering (\cref{Clustering}). During the mask generation stage, We create disjoint masks that cover the entire image, transforming the semantic segmentation task into mask classification. In the feature extraction stage, features are extracted from the generated masks. Lastly, during the clustering stage, cluster labels are assigned to each mask based on the feature. At the same time, we also construct the baseline method.\par

We introduce the Neighborhood Relations-Guided Mask Clustering Algorithm (NeRG-MaskCA) to tackle the challenge of discrete semantics. We describe it in \cref{NeRG-MaskCA}.\par

\subsection{Mask Generation} \label{Mask Generation}
In our framework, the mask proposals generated from the input image \( I \), represented as \( M = \{m_1, m_2, ..., m_n\} \), are designed to be non-overlapping, which can be formally expressed as \( m_i \cap m_j = \emptyset \) for all \( i \neq j \). We can use two distinct strategies to effectively create these proposals: The appearance-based methods and model-based methods.

\textbf{Appearance-based Methods.} In appearance-based methods, such as SLIC ~\cite{achanta2010slic}, it generates mask proposals by taking into account low-level visual features, such as brightness, color, texture, and local gradients.

\textbf{Large-scale Model-based Methods.} In large-scale model-based methods, we employ models like the Segment Anything Model (SAM) ~\cite{SAM} for segmentation tasks. SAM excels in extracting structural information from images, and its zero-shot learning capability allows it to generalize across various image types and tasks. However, the masks generated by SAM alone cannot cover the entire image. Therefore, we treat each connected region ignored by SAM as a separate mask, ensuring comprehensive segmentation of the image.

\begin{figure*}[t]
\centering
\includegraphics[scale=0.6]{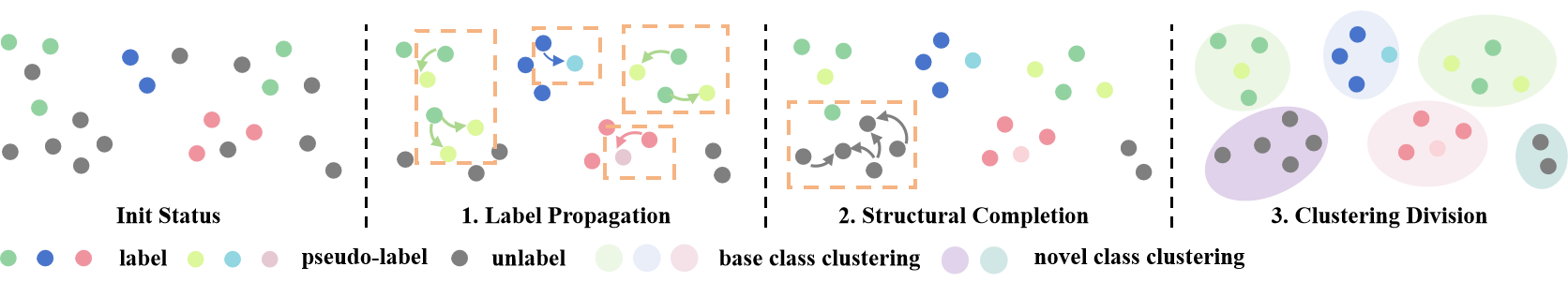}
\caption{\textbf{NeRG-MaskCA.} NeRG-MaskCA is a novel approach comprising three steps: label propagation, structural completion, and clustering division. It starts by assigning pseudo-labels to unlabeled masks based on neighboring labels, identifies high-confidence novel class masks(the rest unlabeled masks), then eliminates the labels of masks in the novel class masks neighborhood, ensuring that novel class masks retain a clustered structure in the feature space, and finally clusters novel class masks.}
\vspace{-0.1in}
\end{figure*} 

\subsection{Feature Extraction} \label{Feature Extraction}

In the feature extraction stage, we start with a set of masks \( M = \{m_1, m_2, ..., m_n\} \). Each mask \( m_i \) is padded and resized. Then, using a feature extractor $f( \cdot )$, we extract a set of features \( F = \{f_1, f_2, ..., f_n\} \) corresponding to each completed image. Existing feature extraction tools for large models have strong generalization ability. We introduce our feature extractor with three typical models.

\textbf{Mask Segmentation Features} A common method is to use the mean of the regional features from the mask generator's feature maps as the mask's features. However, the SAM model primarily focuses on structural information, resulting in less discriminative features for this purpose.

\textbf{Large-scale vision and language Model} This powerful pre-trained vision-language model, such as CLIP ~\cite{CLIP} and OVSeg ~\cite{liang2023open}, demonstrates impressive performance in associating visual and textual concepts. By aligning semantics and images, it naturally achieves a semantic image clustering effect.

\textbf{Large-scale Vision Model} This self-supervised learning method trains deep neural networks without the need for labeled data. The model, such as DINO\cite{DINO,oquab2023dinov2}, has shown remarkable results in various computer vision tasks, such as image classification, object detection, and segmentation, even surpassing some supervised learning methods.

Note that the shape of the masks is irregular, which can be significantly different from the input expected by the feature extractor. Therefore, we pad the mask with the mean value of the rectangle boundary, which is a common padding strategy. Then, we resize the padding mask and input it to the feature extractor. 

\subsection{Generalized Category Discovery Clustering} \label{Clustering}

In the clustering stage, we assign labels to each mask in the set \( M = \{m_1, m_2, ..., m_n\} \) by corresponding features \( F = \{f_1, f_2, ..., f_n\} \), resulting in a corresponding set of labels \( \{l_1, l_2, ..., l_n\} \). These labels are then merged within the same image to form a complete segmentation map. The final segmentation map represents as \( \sum_{i=1}^{n} m_i \times l_i \). \par

We implement a semi-supervised clustering method, a constrained version of the k-means++ clustering algorithm, in the baseline. We establish initial base centroids for the labeled dataset $D_l$ using ground-truth labels and derive novel centroids for the unlabeled dataset $D_u$ (representing novel classes) via the k-means++ algorithm, all the while ensuring these novel centroids are distinct from those of  $D_l$.\par

Throughout each iteration of centroid updates and cluster allocations, each instance in  $D_u$ is eligible for any cluster, with the assignment being based on proximity to centroids and the mask's dimensions. The process finishes when the semi-supervised k-means algorithm stabilizes, at which point each instance in $D_u$ is definitively labeled.\par

During the clustering, the mask set $M$ often contains numerous small masks that lack distinct features, making them difficult to classify. To address this problem, we adopt a nearest-neighbor filling strategy to classify small masks.\par 

\subsection{NeRG-MaskCA} \label{NeRG-MaskCA}
However, the baseline built upon our framework did not adequately address the GCDSS problem. This is primarily due to the introduction of mask generation. It inevitably leads to the challenge of discrete semantics, where a complete concept is divided into several masks with lower-level semantics. Therefore, to address the challenge, we present the Neighborhood Relations-Guided Mask Clustering Algorithm (NeRG-MaskCA). This approach includes three steps: label propagation, structural completion, and clustering division. Initially, NeRG-MaskCA allocates pseudo-labels to unlabeled masks by analyzing adjacent mask labels within the feature space, identifying not annotated masks as high-confidence novel class masks. Subsequently, it eliminates labels from masks near these novel class masks to maintain their distinct clustering. Finally, it uses the clustering algorithm for clustering these novel class masks. See \cref{algorithm} for the algorithm flow.\par

\subsubsection{Label Propagation}

In NeRG-MaskCA's first step, pseudo-labels $l$ are assigned to unlabeled masks by analyzing the labels of neighboring masks in the feature space. We sample the $k$ nearest masks, whose pseudo-label formula is as follows.

\begin{equation}
l = 
\begin{cases} 
\underset{c}{\mathrm{argmax}} \left( \sum_{i=1}^{k} p_i \cdot \mathbb{1}_{\{label_i = c\}} \right)\\
\qquad\qquad ,\text{if } \max \left( \sum_{i=1}^{k} p_i \cdot \mathbb{1}_{\{label_i = c\}} \right) > \theta \\
unlabel,\text{otherwise},
\end{cases} 
\end{equation} 

where  $\theta$ is a lower bound on the confidence we accept and $\mathbb{1}_{equal}$ is an indicator function: 1 if equal is true, 0 if false. Additionally, $p$ represents the confidence of the sample. Initially, masks with labels are assigned a confidence of 1, while unlabeled masks start with a confidence of 0. The confidence of unlabeled samples $p$ is updated as the loop progresses, with the update formula as follows:

\begin{equation}
p = 
\begin{cases} 
\left( \sum_{i=1}^{k} p_i \cdot \mathbb{1}_{\{label_i = c\}} \right)\\
\qquad\qquad ,\text{if } \max \left( \sum_{i=1}^{k} p_i \cdot \mathbb{1}_{\{label_i = c\}} \right) > \theta \\
0,\text{otherwise}.
\end{cases} 
\end{equation}

Then, we will continuously refine the pseudo-labels $l$ and mask confidence $p$ through iterative updates until they converge. Subsequently, unlabeled masks will be confidently identified as belonging to novel classes.

\subsubsection{Structural Completion}  
In the second step, we aim to reinforce the structural integrity of these newly identified novel classes. It is achieved by eliminating the labels of masks in the proximity of these novel class masks, which is a critical process to ensure that these novel classes maintain a distinct, clustered structure within the feature space. The elimination process can be expressed as follows:

\begin{equation}
l = unlabel,\text{if } \left( \sum_{i=1}^{k} \cdot \mathbb{1}_{\{label_i = unlabel\}} \right) > \theta,
\end{equation}
where the parameters are the same as in the previous step.

\subsubsection{Clustering Division}

The final step in our method involves the clustering of the novel class masks. The method is similar to that described in \cref{Clustering}. We utilize a constrained weighted k-means++ clustering algorithm. In this step, the initial clustering centers for the novel classes are deliberately set to be distant from the prototypes of base classes. We focus exclusively on clustering the novel classes. Pseudo-labels are directly adopted for the base classes in the previous step.

\begin{algorithm}
\small
\caption{NeRG-MaskCA}
\label{algorithm}
\begin{algorithmic}[1]
\State \textbf{Input:} $M_u$, $M_l$, $F$,$W$,$L(M_l)$, where  $M_u \cup M_l = M$
\State \textbf{Output:} $L(M_u)$
\State $p(m_u) \gets 0$ for $m_u \in M_u$ , $p(m_l) ~\gets 1$ for $m_l \in M_l$ \Comment{\textbf{Init}}
\For{$x_u \sim M_u$}
    \For{$m' \in M_u \cup M_l$}
        \State dis($m_u$, $m'$) $\gets ||F(m_u) - F(m')||_2$
    \EndFor
    \State find and save top-k nearest mask of $x_u$
\EndFor
\For{$iter \gets 1$ to $max\_iterations$} \Comment{\textbf{Label Propagation}}
    \For{$m_u \in M_u$}
        \State compute label probability distribution (neighbors) 
        \State with weight $p$
        \If{most frequent label $> \theta$}
            \State assign pseudo-label $L(m_u)$ 
            \State update confidence $p(m_u)$
        \EndIf
    \EndFor
\EndFor
\State $p(m_u = unlabel) \gets 1$ \Comment{\textbf{Structural Completion}}
\For{$m_u \in M_u$} 
    \If{weighted probability (neighbors) of unlabel $> \theta$}
        \State $L(m_u) \gets unlabel$
    \EndIf
\EndFor
\State Kmeans++ ( F($m_u$=unlabel), weight=W($m_u$=unlabel) ) 
\State L($m_u$=unlabel) $\gets$ Kmeans++.labels \Comment{\textbf{Clustering Division }}
\State \Return  $L(M_u)$
\end{algorithmic}
\end{algorithm}

\begin{table}[ht]
\centering
\setlength{\tabcolsep}{4pt}
\begin{tabular}{clc}
\toprule
\multirow{2}{*}{\textbf{Comb.}} & \multirow{2}{*}{\textbf{Novel Classes}} & \textbf{Num / Pixel Area} \\
& & \textbf{in Unlabel Set}\\
\midrule
1 & Rider, Truck, Bus, Train                & 1816 / 1.31\% \\
2 & Rider, Bus, Train, Motor.           & 1805 / 1.05\% \\
3 & Wall, Truck, Bus, Train                 & 1767 / 2.08\% \\
4 & Wall, Bus, Train, Motor.            & 1876 / 1.82\% \\
5 & Fence, Truck, Bus, Train                & 1986 / 2.38\% \\
\bottomrule
\end{tabular}
\caption{\textbf{Cityscapes-GCD.} Our dataset includes five combinations, each with a labeled set (1390 images) and an unlabeled set (2085 images). It features 15 base classes and 4 novel classes. We also provide detailed information on the novel classes in the unlabeled set, including image number (Num) and pixel area proportion (Pixel Area).}
\label{dataset_info}
\end{table}

\section{Experiment}
\subsection{Experimental Setup}

\textbf{Dateset.} We introduce a new dataset, Cityscapes-GCD, to address the problem of Generalized Category Discovery in Semantic Segmentation. It is built upon the Cityscapes dataset~\cite{Cityscapes}. Cityscapes-GCD is divided into two subsets: the labeled set $D_{l}$ and the unlabeled set $D_{u}$. The labeled set $D_{l}$ contains only the base classes, while the unlabeled set $D_{u}$ includes both the base classes and novel classes. In Cityscapes-GCD, we evaluate the robustness and generalization capabilities of our proposed method using multiple combinations of novel classes. Each combination contains 15 base classes and 4 novel classes. Details of the dataset splits, and class distributions are provided in \cref{dataset_info}. By evaluating our method on various combinations of novel classes, it demonstrates its effectiveness for discovering and segmenting novel classes in unlabeled data.\par

\begin{table*}[ht]
\centering
\begin{tabular}{ccccccc}
\hline
\textbf{Combination} & \multicolumn{3}{c}{\textbf{Baseline}} & \multicolumn{3}{c}{\textbf{NeRG-MaskCA}} \\
\cline{2-7}
& \textbf{Base Class} & \textbf{Novel Class} & \textbf{Avg Class} & \textbf{Base Class} & \textbf{Novel Class} & \textbf{Avg Class}\\
\hline
Comb. 1 & 31.99  & 3.38 & 25.97 & 46.12 & 30.61 & 42.86 \\
Comb. 2 & 28.38 & 2.36 & 22.9 & 46.62 & 28.94 & 42.90\\
Comb. 3 & 31.01 & 2.10 & 24.92 & 46.42 & 30.74 & 43.12\\
Comb. 4 & 32.3 & 3.86 & 26.31 & 46.84 & 28.03 & 42.88 \\
Comb. 5 & 28.91 & 5.88 & 24.06 & 45.65 & 33.18 & 43.02 \\
\hline
\textbf{Average mIoU} & 30.52 & 3.52 & 24.83 & \textbf{46.33} & \textbf{30.30} & \textbf{42.96}\\
\hline
\end{tabular}
\caption{\textbf{Comparison of the baseline and NeRG-MaskCA across five class combinations.} NeRG-MaskCA outperforms the baseline compared to the five class combinations.}
\label{miou_comparison}
\end{table*}

\textbf{Metric.} In previous works on Generalized Category Discovery (GCD), Hungarian matching has been applied to assigning both base and novel classes. However, this approach may lead to hybrid classes consisting of both base and novel classes being greedily matched to novel classes, while the corresponding base class goes undiscovered. It is unreasonable, as we would not consider this a true discovery of a novel class, but rather a confusion with an existing base class. To rectify the problem, we introduce a refined evaluation metric that imposes stringent criteria on the classification capabilities of the model. For the initial $k_{base}$ base classes, the model is expected to produce precise labels. For the novel classes, we allow the number of predicted novel class data $k_{pred}$ to be unequal to $k_{novel}$. We apply Hungarian matching to identify the peak mIoU for up to $k_{novel}$ novel classes. Any additional predicted classes (where $k_{pred} > k_{novel}$) are considered incorrect predictions. We measure the model performance using the mean Intersection-over-Union (\text{mIoU}).

\textbf{Implementation Details.} Our experimental framework is implemented using Pytorch, harnessing the power of an NVIDIA RTX 2080Ti GPU. We adopt SAM for mask generation. The configuration parameters of SAM are guided by~\cite{chen2023semantic}: $points\_per\_side$ is set to 32, $pred\_iou\_thresh$ at 0.86, $stability\_score\_thresh$ at 0.92. Smaller masks are given semantic precedence when masks are overlapping. The remaining pixels with contiguous regions are considered as a unified mask. For the feature extractor, we adopt DINO v2 for feature extraction. For NeRG-MaskCA: $\theta$ is set to 0.1 and $k$ is set to 10. We iterate the dilation step 10 times to achieve convergence and precise label allocation.

\section{Comparison with Baseline}

We compare our NeRG-MaskCA with the baseline of our framework in the Cityscapes-GCD dataset. \Cref{miou_comparison} delineates the results, underlining the strengths of our approach. This indicates that our method outperforms the baseline significantly. Performance metrics, evaluated across diverse class combinations, show marked enhancements in discovering novel classes. 

\section{Ablation Study}

To better investigate the effectiveness of NeRG-MaskCA and the different components of NeRG-MaskCA, we conduct ablation studies, shown in \cref{ablation}.

\begin{table}[ht]
\centering
\begin{tabular}{cccc}
\hline
\textbf{Clustering} & \textbf{Label} & \textbf{Struct} & \textbf{mIoU (\%)} \\
\textbf{Div.} & \textbf{Prop.} & \textbf{Comp.} & \textbf{(Base / Novel / Avg)} \\
\hline
\checkmark & - & - & 30.52 / ~3.52 / 24.83 \\
\checkmark & \checkmark & - & 46.31 / 23.92 / 41.60 \\
\checkmark & \checkmark & \checkmark & \textbf{46.33} / \textbf{30.30} / \textbf{42.96} \\
\hline
\end{tabular}
\caption{\textbf{Comparison of components.}}
\label{ablation}
\end{table}

\subsection{Parameters Study}
\begin{figure}[t]
\centering
\includegraphics[scale=0.28]{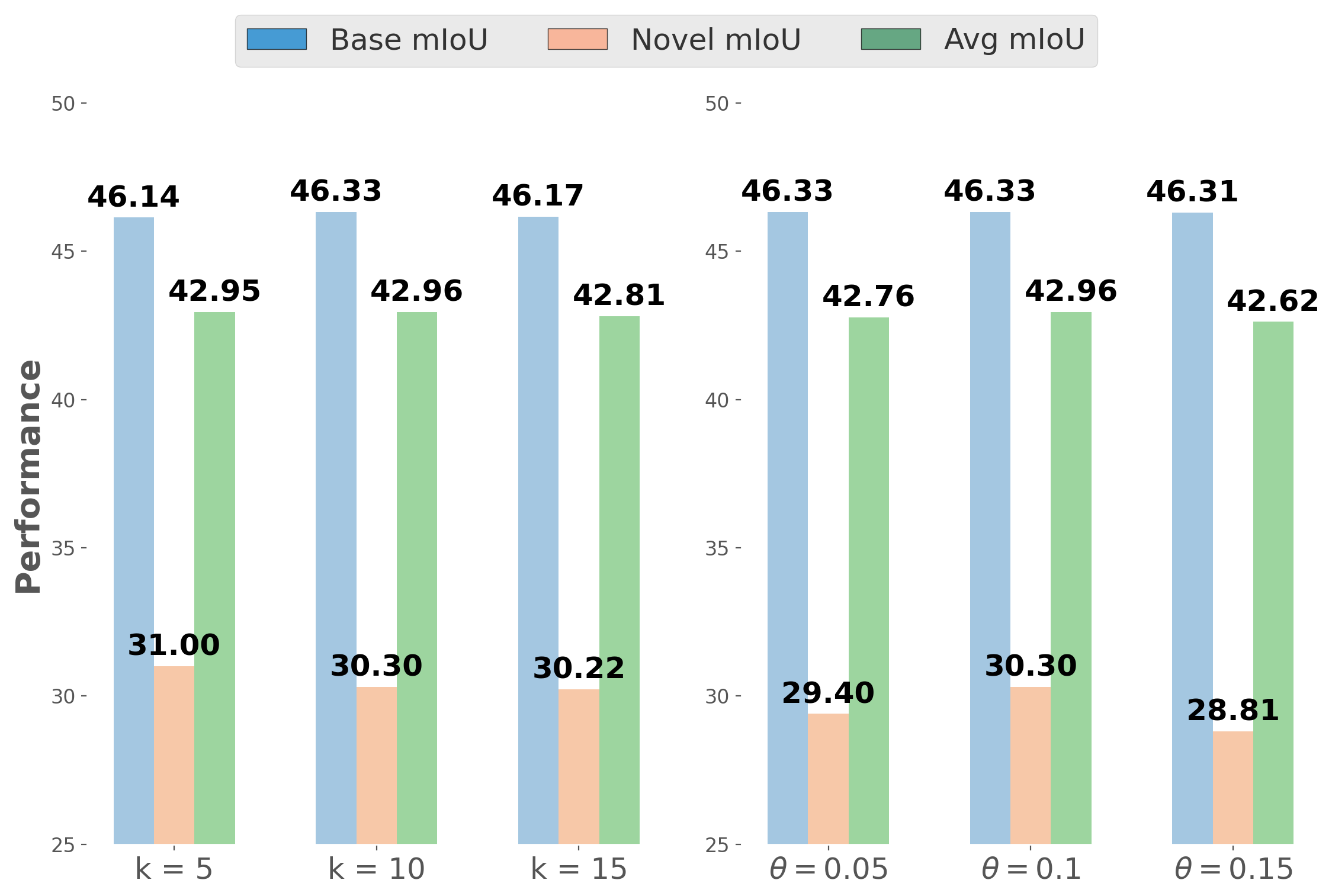}
\caption{\textbf{Parameter analysis of $k$ and $\theta$.} The nearest mask number $k$ varies among 5, 10, and 15. The lower bound confidence $\theta$ for pseudo-label changes among 0.05, 0.10, and 0.15. The performance of our approach is relatively stable.}
\label{ab}
\vspace{-0.1in}
\end{figure}

\Cref{ab} presents the parameters study of the nearest mask number $k$ and the lower bound confidence $\theta$ for pseudo-label. As  $k$ increases, the quantity of neighbor masks increases, but the quality decreases. As $\theta$ increases, more masks are introduced as novel classes, while the average quality of novel classes samples decreases. Both of the parameters represent a trade-off between quantity and quality.

\subsection{Mask Generation Approach}
We conduct a comparison to assess the mask generation capabilities of SLIC and SAM, as shown in \cref{mask_generation}. It is observed that SAM significantly outperforms SLIC.

\begin{table}[ht]
\centering
\begin{tabular}{lcc}
\hline
\multirow{2}{*}{\textbf{Metric}} & \textbf{SLIC} & \textbf{SAM} \\
& \textbf{(Base / Novel / Avg)} & \textbf{(Base / Novel / Avg)} \\
\hline
mIoU & 16.60 / 9.03 / 15.00 &  \textbf{46.33} / \textbf{30.30} / \textbf{42.96}\\
\hline
\end{tabular}
\caption{\textbf{Comparison of performance between SLIC and SAM.}}
\label{mask_generation}
\end{table}

\begin{figure*}[t]
\vspace{-0.2in}
\centering
\includegraphics[scale=0.6]{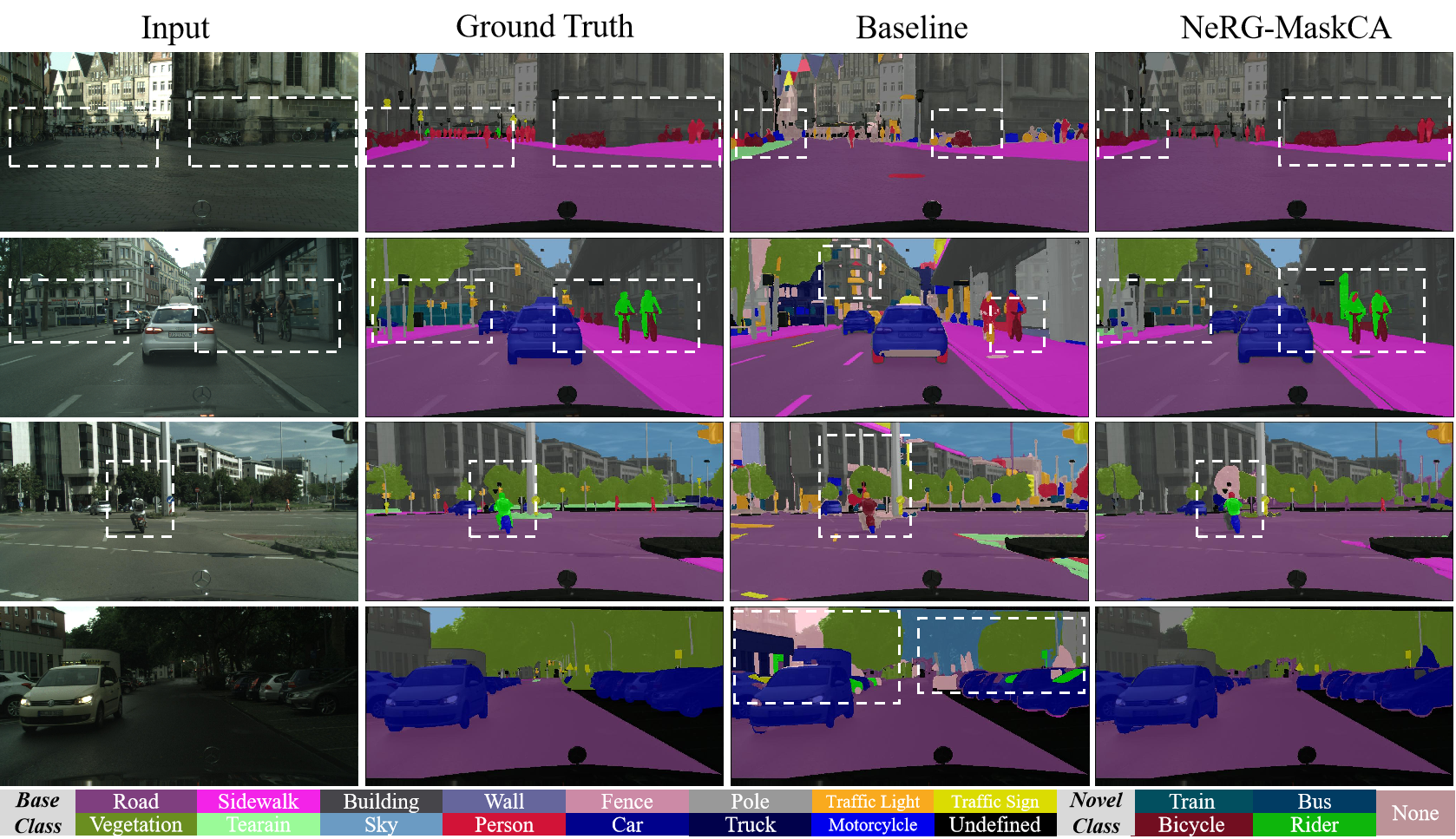}
\caption{\textbf{Visualization comparison of our approach with baseline in Cityscapes-GCD dataset.} The white boxes indicate the actual location of the novel classes or where the method predicted novel classes. Rows 1-3 actually contain novel classes, and the performance of NeRG-MaskCA in predicting novel classes is notably superior to the baseline. Row 4 depicts images without novel classes and the prediction of our approach contains no novel class, but the baseline incorrectly predicts novel classes.}
\vspace{-0.1in}
\end{figure*} 

\subsection{Feature Extraction Methods}
In our analysis of feature extraction methods (See \cref{table:feature_extraction_models}), features extracted by SAM emphasize structural elements, potentially sacrificing semantic details. Large-scale vision and language models, such as CLIP and OVSeg, perform well in semantics but face challenges in single-modality scenarios without textual cues. Conversely, DINO v1 and v2 that trained contrastively are proved more suitable to extract feature of masks by leveraging their pre-training task.

\begin{table}[ht]
\centering
\begin{tabular}{lcccc}
\hline
\textbf{Model} & \textbf{Base Class} & \textbf{Novel Class} & \textbf{Avg Class} \\
\hline
SAM~\cite{SAM}  & 15.38 & ~0.95 & 12.34 \\
CLIP~\cite{CLIP} & 17.98 & ~0.15 & 14.23 \\
OVSEG~\cite{OVSeg}  & 42.70 & 11.77 & 36.18 \\
DINO v1~\cite{DINO} & 43.51 & 11.40 & 36.74 \\
DINO v2~\cite{oquab2023dinov2} & \textbf{46.33} & \textbf{30.30} & \textbf{42.96} \\
\hline
\end{tabular}
\caption{\textbf{Comparison of different feature extraction models.}}
\label{table:feature_extraction_models}
\end{table}

\subsection{Self-training}

Our method can generate novel class pseudo-labels, enabling models that are initially unable to segment these novel classes to acquire the capability for such segmentation through self-training~\cite{ST}. \Cref{tab:effect_integration} shows the result of integration into DeepLab-v3+\cite{DeepLabv3+} via self-training. This demonstrates the potential extension prospects of our approach.

\begin{table}[ht]
\centering
\begin{tabular}{lcccc}
\hline
\multirow{2}{*}{\textbf{Approach}} & \multicolumn{3}{c}{\textbf{mIoU (\%)}} \\
& \textbf{Base Class} & \textbf{Novel Class} & \textbf{Avg Class} \\
\hline
ST~\cite{ST}      &  68.21 & 0.00 & 53.85 \\
ST+Ours &  66.46 & 43.56 & \textbf{62.48} \\
\hline
\end{tabular}
\caption{\textbf{Results of integration our approach into DeepLab-v3+ via self-training.}}
\label{tab:effect_integration}
\end{table}

\section{Conclusion}

In this work, we introduce a new setting of Generalized Category Discovery in Semantic Segmentation (GCDSS)  that effectively segments unlabeled images by leveraging prior knowledge from labeled base classes. Unlike NCDSS, GCDSS does not impose the constraint that unlabeled images must contain pixels from novel classes, enhancing its versatility.  We introduce a general framework to tackle this challenge and establish a baseline.  Additionally, we propose the NeRG-MaskCA algorithm to extract new class information efficiently from unlabeled data.  This innovative method paves the way for advancements in generalized category discovery, extending the applicability of semantic segmentation in various real-world scenarios.

{
\small
\bibliographystyle{ieeenat_fullname}
\bibliography{main}

\begin{thebibliography}{43}
\providecommand{\natexlab}[1]{#1}
\providecommand{\url}[1]{\texttt{#1}}
\expandafter\ifx\csname urlstyle\endcsname\relax
  \providecommand{\doi}[1]{doi: #1}\else
  \providecommand{\doi}{doi: \begingroup \urlstyle{rm}\Url}\fi

\bibitem[Achanta et~al.(2010)Achanta, Shaji, Smith, Lucchi, Fua, and S{\"u}sstrunk]{achanta2010slic}
Radhakrishna Achanta, Appu Shaji, Kevin Smith, Aurelien Lucchi, Pascal Fua, and Sabine S{\"u}sstrunk.
\newblock Slic superpixels.
\newblock Technical report, 2010.

\bibitem[Carion et~al.(2020)Carion, Massa, Synnaeve, Usunier, Kirillov, and Zagoruyko]{DETR}
Nicolas Carion, Francisco Massa, Gabriel Synnaeve, Nicolas Usunier, Alexander Kirillov, and Sergey Zagoruyko.
\newblock End-to-end object detection with transformers.
\newblock In \emph{ECCV}, 2020.

\bibitem[Caron et~al.(2021)Caron, Touvron, Misra, J\'egou, Mairal, Bojanowski, and Joulin]{DINO}
Mathilde Caron, Hugo Touvron, Ishan Misra, Herv\'e J\'egou, Julien Mairal, Piotr Bojanowski, and Armand Joulin.
\newblock Emerging properties in self-supervised vision transformers.
\newblock In \emph{ICCV}, 2021.

\bibitem[Cen et~al.(2021)Cen, Yun, Cai, Wang, and Liu]{cen2021deep}
Jun Cen, Peng Yun, Junhao Cai, Michael~Yu Wang, and Ming Liu.
\newblock Deep metric learning for open world semantic segmentation. 2021 ieee.
\newblock In \emph{ICCV}, 2021.

\bibitem[Chen et~al.(2023)Chen, Yang, and Zhang]{chen2023semantic}
Jiaqi Chen, Zeyu Yang, and Li Zhang.
\newblock Semantic segment anything.
\newblock \url{https://github.com/fudan-zvg/Semantic-Segment-Anything}, 2023.

\bibitem[Chen et~al.(2018{\natexlab{a}})Chen, Papandreou, Kokkinos, Murphy, and Yuille]{DeepLab-v2}
Liang{-}Chieh Chen, George Papandreou, Iasonas Kokkinos, Kevin Murphy, and Alan~L. Yuille.
\newblock Deeplab: Semantic image segmentation with deep convolutional nets, atrous convolution, and fully connected crfs.
\newblock \emph{{IEEE} TPAMI}, 2018{\natexlab{a}}.

\bibitem[Chen et~al.(2018{\natexlab{b}})Chen, Zhu, Papandreou, Schroff, and Adam]{DeepLabv3+}
Liang{-}Chieh Chen, Yukun Zhu, George Papandreou, Florian Schroff, and Hartwig Adam.
\newblock Encoder-decoder with atrous separable convolution for semantic image segmentation.
\newblock In \emph{ECCV}, 2018{\natexlab{b}}.

\bibitem[Cheng et~al.(2021)Cheng, Schwing, and Kirillov]{MaskFormer}
Bowen Cheng, Alexander~G. Schwing, and Alexander Kirillov.
\newblock Per-pixel classification is not all you need for semantic segmentation.
\newblock In \emph{NeurIPS}, 2021.

\bibitem[Chi et~al.(2022)Chi, Liu, Yang, Lan, Liu, Han, Niu, Zhou, and Sugiyama]{Meta_Discovery}
Haoang Chi, Feng Liu, Wenjing Yang, Long Lan, Tongliang Liu, Bo Han, Gang Niu, Mingyuan Zhou, and Masashi Sugiyama.
\newblock Meta discovery: Learning to discover novel classes given very limited data.
\newblock In \emph{ICLR}, 2022.

\bibitem[Cordts et~al.(2016)Cordts, Omran, Ramos, Rehfeld, Enzweiler, Benenson, Franke, Roth, and Schiele]{Cityscapes}
Marius Cordts, Mohamed Omran, Sebastian Ramos, Timo Rehfeld, Markus Enzweiler, Rodrigo Benenson, Uwe Franke, Stefan Roth, and Bernt Schiele.
\newblock The cityscapes dataset for semantic urban scene understanding.
\newblock In \emph{CVPR}, 2016.

\bibitem[Dehghani et~al.(2023)Dehghani, Djolonga, Mustafa, Padlewski, Heek, Gilmer, Steiner, Caron, Geirhos, Alabdulmohsin, et~al.]{dehghani2023scaling}
Mostafa Dehghani, Josip Djolonga, Basil Mustafa, Piotr Padlewski, Jonathan Heek, Justin Gilmer, Andreas~Peter Steiner, Mathilde Caron, Robert Geirhos, Ibrahim Alabdulmohsin, et~al.
\newblock Scaling vision transformers to 22 billion parameters.
\newblock In \emph{ICML}, 2023.

\bibitem[Fini et~al.(2021)Fini, Sangineto, Lathuili{\`{e}}re, Zhong, Nabi, and Ricci]{UNO}
Enrico Fini, Enver Sangineto, St{\'{e}}phane Lathuili{\`{e}}re, Zhun Zhong, Moin Nabi, and Elisa Ricci.
\newblock A unified objective for novel class discovery.
\newblock In \emph{ICCV}, 2021.

\bibitem[Ghiasi et~al.(2022)Ghiasi, Gu, Cui, and Lin]{ghiasi2022scaling}
Golnaz Ghiasi, Xiuye Gu, Yin Cui, and Tsung-Yi Lin.
\newblock Scaling open-vocabulary image segmentation with image-level labels.
\newblock In \emph{ECCV}, 2022.

\bibitem[Han et~al.(2019)Han, Vedaldi, and Zisserman]{DTC}
Kai Han, Andrea Vedaldi, and Andrew Zisserman.
\newblock Learning to discover novel visual categories via deep transfer clustering.
\newblock In \emph{ICCV}, 2019.

\bibitem[He et~al.(2017)He, Gkioxari, Doll{\'{a}}r, and Girshick]{MaskRCNN}
Kaiming He, Georgia Gkioxari, Piotr Doll{\'{a}}r, and Ross~B. Girshick.
\newblock Mask {R-CNN}.
\newblock In \emph{ICCV}, 2017.

\bibitem[Hendrycks and Gimpel(2016)]{hendrycks2016baseline}
Dan Hendrycks and Kevin Gimpel.
\newblock A baseline for detecting misclassified and out-of-distribution examples in neural networks.
\newblock \emph{arXiv preprint arXiv:1610.02136}, 2016.

\bibitem[Joseph et~al.(2022)Joseph, Paul, Aggarwal, Biswas, Rai, Han, and Balasubramanian]{NCDwF}
K.~J. Joseph, Sujoy Paul, Gaurav Aggarwal, Soma Biswas, Piyush Rai, Kai Han, and Vineeth~N. Balasubramanian.
\newblock Novel class discovery without forgetting.
\newblock In \emph{ECCV}, 2022.

\bibitem[Kirillov et~al.(2023)Kirillov, Mintun, Ravi, Mao, Rolland, Gustafson, Xiao, Whitehead, Berg, Lo, et~al.]{SAM}
Alexander Kirillov, Eric Mintun, Nikhila Ravi, Hanzi Mao, Chloe Rolland, Laura Gustafson, Tete Xiao, Spencer Whitehead, Alexander~C Berg, Wan-Yen Lo, et~al.
\newblock Segment anything.
\newblock \emph{arXiv preprint arXiv:2304.02643}, 2023.

\bibitem[Li et~al.(2023)Li, Fan, Huo, and Gao]{InterClassNCD}
Wenbin Li, Zhichen Fan, Jing Huo, and Yang Gao.
\newblock Modeling inter-class and intra-class constraints in novel class discovery.
\newblock In \emph{CVPR}, 2023.

\bibitem[Liang et~al.(2023)Liang, Wu, Dai, Li, Zhao, Zhang, Zhang, Vajda, and Marculescu]{OVSeg}
Feng Liang, Bichen Wu, Xiaoliang Dai, Kunpeng Li, Yinan Zhao, Hang Zhang, Peizhao Zhang, Peter Vajda, and Diana Marculescu.
\newblock Open-vocabulary semantic segmentation with mask-adapted clip.
\newblock In \emph{CVPR}, 2023.

\bibitem[Long et~al.(2015)Long, Shelhamer, and Darrell]{FCN}
Jonathan Long, Evan Shelhamer, and Trevor Darrell.
\newblock Fully convolutional networks for semantic segmentation.
\newblock In \emph{CVPR}, 2015.

\bibitem[Luo et~al.(2023)Luo, Bao, Wu, He, and Li]{luo2023segclip}
Huaishao Luo, Junwei Bao, Youzheng Wu, Xiaodong He, and Tianrui Li.
\newblock Segclip: Patch aggregation with learnable centers for open-vocabulary semantic segmentation.
\newblock In \emph{ICML}, 2023.

\bibitem[Oquab et~al.(2023)Oquab, Darcet, Moutakanni, Vo, Szafraniec, Khalidov, Fernandez, Haziza, Massa, El-Nouby, et~al.]{oquab2023dinov2}
Maxime Oquab, Timoth{\'e}e Darcet, Th{\'e}o Moutakanni, Huy Vo, Marc Szafraniec, Vasil Khalidov, Pierre Fernandez, Daniel Haziza, Francisco Massa, Alaaeldin El-Nouby, et~al.
\newblock Dinov2: Learning robust visual features without supervision.
\newblock \emph{arXiv preprint arXiv:2304.07193}, 2023.

\bibitem[Ouldnoughi et~al.(2023)Ouldnoughi, Kuo, and Kira]{CLIP-GCD}
Rabah Ouldnoughi, Chia{-}Wen Kuo, and Zsolt Kira.
\newblock Clip-gcd: Simple language guided generalized category discovery.
\newblock \emph{arXiv preprint arXiv:2305.10420}, 2023.

\bibitem[Pu et~al.(2023)Pu, Zhong, and Sebe]{DCCL}
Nan Pu, Zhun Zhong, and Nicu Sebe.
\newblock Dynamic conceptional contrastive learning for generalized category discovery.
\newblock In \emph{CVPR}, 2023.

\bibitem[Radford et~al.(2021)Radford, Kim, Hallacy, Ramesh, Goh, Agarwal, Sastry, Askell, Mishkin, Clark, et~al.]{CLIP}
Alec Radford, Jong~Wook Kim, Chris Hallacy, Aditya Ramesh, Gabriel Goh, Sandhini Agarwal, Girish Sastry, Amanda Askell, Pamela Mishkin, Jack Clark, et~al.
\newblock Learning transferable visual models from natural language supervision.
\newblock In \emph{ICML}, 2021.

\bibitem[Rao et~al.(2022)Rao, Zhao, Chen, Tang, Zhu, Huang, Zhou, and Lu]{rao2022denseclip}
Yongming Rao, Wenliang Zhao, Guangyi Chen, Yansong Tang, Zheng Zhu, Guan Huang, Jie Zhou, and Jiwen Lu.
\newblock Denseclip: Language-guided dense prediction with context-aware prompting.
\newblock In \emph{CVPR}, 2022.

\bibitem[Riz et~al.(2023)Riz, Saltori, Ricci, and Poiesi]{NCD3DPoint}
Luigi Riz, Cristiano Saltori, Elisa Ricci, and Fabio Poiesi.
\newblock Novel class discovery for 3d point cloud semantic segmentation.
\newblock In \emph{CVPR}, 2023.

\bibitem[Vaswani et~al.(2017)Vaswani, Shazeer, Parmar, Uszkoreit, Jones, Gomez, Kaiser, and Polosukhin]{Transformer}
Ashish Vaswani, Noam Shazeer, Niki Parmar, Jakob Uszkoreit, Llion Jones, Aidan~N. Gomez, Lukasz Kaiser, and Illia Polosukhin.
\newblock Attention is all you need.
\newblock In \emph{NeurIPS}, 2017.

\bibitem[Vaze et~al.(2022)Vaze, Han, Vedaldi, and Zisserman]{GCD}
Sagar Vaze, Kai Han, Andrea Vedaldi, and Andrew Zisserman.
\newblock Generalized category discovery.
\newblock In \emph{CVPR}, 2022.

\bibitem[Xia et~al.(2020)Xia, Zhang, Liu, Shen, and Yuille]{xia2020synthesize}
Yingda Xia, Yi Zhang, Fengze Liu, Wei Shen, and Alan~L Yuille.
\newblock Synthesize then compare: Detecting failures and anomalies for semantic segmentation.
\newblock In \emph{ECCV}, 2020.

\bibitem[Xin~Wen and Qi(2023)]{Parametric_GCD}
Bingchen~Zhao Xin~Wen and Xiaojuan Qi.
\newblock Parametric classification for generalized category discovery: A baseline study.
\newblock In \emph{ICCV}, 2023.

\bibitem[Xu et~al.(2022)Xu, Zhang, Wei, Lin, Cao, Hu, and Bai]{xu2022simple}
Mengde Xu, Zheng Zhang, Fangyun Wei, Yutong Lin, Yue Cao, Han Hu, and Xiang Bai.
\newblock A simple baseline for open-vocabulary semantic segmentation with pre-trained vision-language model.
\newblock In \emph{ECCV}, 2022.

\bibitem[Yanan~Wu and Feng(2023)]{MetaGCD}
Yang~Wang Yanan~Wu, Zhixiang~Chi and Songhe Feng.
\newblock Metagcd: Learning to continually learn in generalized category discovery.
\newblock In \emph{ICCV}, 2023.

\bibitem[Yang et~al.(2022)Yang, Zhuo, Qi, Shi, and Gao]{ST}
Lihe Yang, Wei Zhuo, Lei Qi, Yinghuan Shi, and Yang Gao.
\newblock {ST++:} make self-training work better for semi-supervised semantic segmentation.
\newblock In \emph{CVPR}, 2022.

\bibitem[Yang et~al.(2023)Yang, Wang, Deng, and Zhang]{Bootstrap_Your_Own_Prior}
Muli Yang, Liancheng Wang, Cheng Deng, and Hanwang Zhang.
\newblock Bootstrap your own prior: Towards distribution-agnostic novel class discovery.
\newblock In \emph{CVPR}, 2023.

\bibitem[Zhao and Aodha(2023)]{IGCD}
Bingchen Zhao and Oisin~Mac Aodha.
\newblock Incremental generalized category discovery.
\newblock \emph{arXiv preprint arXiv:2304.14310}, 2023.

\bibitem[Zhao et~al.(2023)Zhao, Wen, and Han]{Learning_Semi-supervised_GCD}
Bingchen Zhao, Xin Wen, and Kai Han.
\newblock Learning semi-supervised gaussian mixture models for generalized category discovery.
\newblock \emph{arXiv preprint arXiv:2305.06144}, 2023.

\bibitem[Zhao et~al.(2017)Zhao, Shi, Qi, Wang, and Jia]{PSPNet}
Hengshuang Zhao, Jianping Shi, Xiaojuan Qi, Xiaogang Wang, and Jiaya Jia.
\newblock Pyramid scene parsing network.
\newblock In \emph{CVPR}, 2017.

\bibitem[Zhao et~al.(2022)Zhao, Zhong, Sebe, and Lee]{NCDSS}
Yuyang Zhao, Zhun Zhong, Nicu Sebe, and Gim~Hee Lee.
\newblock Novel class discovery in semantic segmentation.
\newblock In \emph{CVPR}, 2022.

\bibitem[Zheng et~al.(2021)Zheng, Lu, Zhao, Zhu, Luo, Wang, Fu, Feng, Xiang, Torr, and Zhang]{SETR}
Sixiao Zheng, Jiachen Lu, Hengshuang Zhao, Xiatian Zhu, Zekun Luo, Yabiao Wang, Yanwei Fu, Jianfeng Feng, Tao Xiang, Philip H.~S. Torr, and Li Zhang.
\newblock Rethinking semantic segmentation from a sequence-to-sequence perspective with transformers.
\newblock In \emph{CVPR}, 2021.

\bibitem[Zhong et~al.(2021{\natexlab{a}})Zhong, Fini, Roy, Luo, Ricci, and Sebe]{NCL}
Zhun Zhong, Enrico Fini, Subhankar Roy, Zhiming Luo, Elisa Ricci, and Nicu Sebe.
\newblock Neighborhood contrastive learning for novel class discovery.
\newblock In \emph{CVPR}, 2021{\natexlab{a}}.

\bibitem[Zhong et~al.(2021{\natexlab{b}})Zhong, Zhu, Luo, Li, Yang, and Sebe]{OpenMix}
Zhun Zhong, Linchao Zhu, Zhiming Luo, Shaozi Li, Yi Yang, and Nicu Sebe.
\newblock Openmix: Reviving known knowledge for discovering novel visual categories in an open world.
\newblock In \emph{CVPR}, 2021{\natexlab{b}}.

\end{thebibliography}
}

\end{document}